# Algorithms for Irrelevance-Based Partial MAPs

Solomon E. Shimony
Computer Science Department
Box 1910, Brown University
Providence, RI 02912
ses@cs.brown.edu

## Abstract

Irrelevance-based partial MAPs are useful constructs for domain-independent explanation using belief networks. We look at two definitions for such partial MAPs, and prove important properties that are useful in designing algorithms for computing them effectively. We make use of these properties in modifying our standard MAP best-first algorithm, so as to handle irrelevance-based partial MAPs.

## 1 INTRODUCTION

Probabilistic explanation, finding causes for observed facts (or evidence), is an extremely important aspect of Artificial Intelligence in general, and probabilistic reasoning in particular. For example, [Charniak and Goldman, 1988], views the understanding of stories as finding high probability facts given the evidence as an explanation of the natural language input text. In automated medical diagnosis (for example the work of [Cooper, 1984], and [Peng and Reggia, 1987]), one wants to find the disease or set of diseases of highest probability given the observed symptoms. In vision processing, recent research formulates the problem in terms of finding some set of objects that have the highest probability given the evidence (the image).

There is, however, no agreement on what should be maximized in finding a good explanation. In fact, Poole discussed six different schemes of probabilistic explanation in [Poole and Provan, 1990], and even these are not exhaustive. One of the schemes discussed was Maximum A-Posteriori models (MAP), which was presented in [Pearl, 1988] by the name of Maximum Probability Explanation (MPE). A MAP is a maximum probability (given the evidence) assignment to *all* the variables. We call such assignments *complete* MAPs, as opposed to *partial* MAPs which are maximum probability assignments to *some* of the variables. MAPs are useful for finding best globally consistent explanations, as argued by Pearl. In [Charniak and Shimony, 1990], we showed that MAPs are useful for explanation by demonstrating that MAP explanations are equivalent to complete assignment cost-based abduction. Cost based abduction is a variant of Hobbs' and Stickel's weighted abduction (see [Hobbs and Stickel, 1988]), which they used for natural language story understanding.

In the rest of this section, we will present the essence of earlier papers: [Shimony and Charniak, 1990] (an algorithm for complete MAPs), and [Shimony, 1991] (definitions of irrelevance-based assignments). The following sections will deal with how we modify the algorithm for complete MAPs to handle partial MAPs. We assume here, as well as in related papers, that the world knowledge is represented as a belief network (Bayesian network).

In [Shimony, 1991], we proposed a new, domain-independent method of highest likelihood explanations called *irrelevance-based (partial) MAPs*. The idea is that the standard MAP solution, that of finding the most-probable complete model given the evidence, suffers from the *overspecification problem* (an instance of which appears in [Pearl, 1988], and [Shimony, 1991]). Our solution is a generalization of Pearl's idea of "circumscribing explanations". Pearl claimed that there is no need to consider the assignment to nodes which have no evidence coming in from below (evidential support).

In many cases the evidential support is insufficient as a criterion for deciding which nodes are irrelevant, as shown in [Shimony, 1991]. In that paper, we defined irrelevance-based assignments as assignments where every unassigned node is irrelevant. We then defined our notion of explanation, irrelevance-based MAP, as the irrelevance-based partial assignment of highest probability (given the evidence).

## 1.1 IRRELEVANCE-BASED ASSIGNMENTS

We proceeded to give irrelevance a more formal footing, using statistical independence as a criterion for irrelevance. There were two such definitions of irrelevance-based assignments: independence-based partial assignments, and $\delta$-independence based partial assignments; the second being a more general concept that was introduced because independence-based assignments were too restrictive and captured the intuitive meaning of irrelevance only in special cases.

We introduced the notion of independence given a (partial) assignment[1] $\mathcal{A}^S$. We use the notation $In(a, b|\mathcal{A}^S)$ to mean that $a$ and $b$ are independent given an *assignment* $\mathcal{A}^S$, where $a$ and $b$ are either assignments (assignments are used interchangeably with sample-space events), or sets of nodes. Equivalently, we can say that $P(a|\mathcal{A}^S) = P(a|b, \mathcal{A}^S)$. The latter constraint actually is a set of simple constraints, one for each possible assignments to the nodes of $a$ and $b$. This is similar to Pearl's notation of $I(a, S, b)$ stating that $a$ and $b$ are independent given $S$, where $a, S, b$ are sets of variables. The difference is that the latter implies the former, but not vice versa. That is because our notion only states that independence occurs given a *particular* assignment to $S$, whereas Pearl's notion states that independence occurs given *any* assignment to $S$. That is, $I(a, S, b)$ stands for a set-wise larger set of constraints than $In(a, b|\mathcal{A}^S)$.

An assignment can be seen as a set of pairs, where a pair $(v, V)$ means that node $v$ is assigned the value $V$. The function $nodes(\mathcal{A}^S)$ evaluates to the set of nodes assigned (in this case $S$). We say that assignment $\mathcal{A}$ *subsumes* assignment $\mathcal{B}$ iff $\mathcal{A} \subseteq \mathcal{B}$. The evidence $\mathcal{E}$ is assumed to be an assignment. We say that an assignment $\mathcal{A}^S$ is *evidentially supported* by $\mathcal{E}$ iff every node $v \in S$ is either an evidence node or there exists a path form $v$ to some evidence node. Likewise, $\mathcal{A}^S$ is properly evidentially supported by $\mathcal{E}$ iff every node $v \in S$ is either an evidence node or there exists a path form $v$ to an evidence node that traverses only nodes in $S$.

Our definition of irrelevance-based assignments relies on the directionality of belief networks, the "cause and effect" directionality. The potential causes of a node $v$ are its parents, $\uparrow(v)$, and we do not assign (i.e. are not interested in) variables that are irrelevant to the evidence given the causes. We defined our first notion of an irrelevance-based partial assignment formally (and called it *independence-based partial assignment*):

**Definition 1** *An assignment $\mathcal{A}^S$ is an* independence-based *assignment (IB assignment for short) iff for every node $v \in S$, $\mathcal{A}^{\{v\}}$ is independent of all its ancestors that are not in $S$, given $\mathcal{A}^{S\uparrow(v)}$.* [2]

If $v$ is independent of its unassigned parents, as in definition 1, we say that the IB constraint holds at $v$. The idea behind this definition is that the unassigned nodes above each assigned node $v$ should remain unassigned if they cannot affect $v$ (and cannot be used to explain $v$). Nodes that are not above $v$ are never used as an explanation of $v$ anyway, as we stated implicitly earlier.

**Definition 2** *An IB assignment $\mathcal{A}^S$ is an independence based MAP (IB-MAP) w.r.t. to evidence $\mathcal{E}$ iff $\mathcal{A}^S$ is evidentially supported and subsumed by $\mathcal{E}$, and there is no other IB assignment assignment evidentially supported and subsumed by $\mathcal{E}$ of greater probability given the evidence.*

Clearly, since $\mathcal{A}^S$ is subsumed by $\mathcal{E}$, then $P(\mathcal{E}|\mathcal{A}^S) = 1$, whenever $P(\mathcal{A}^S) \neq 0$. In fact, we are only interested in MAPs that are maximal w.r.t. subsumption, because they assign fewer variables and thus lead to "simpler" explanations. This distinction is immaterial if the distribution of the belief network is strictly positive, because then if assignment $\mathcal{A}$ subsumes $\mathcal{B}$ (with $\mathcal{A} \neq \mathcal{B}$) then it also has a strictly greater probability.

We need to maximize $P(\mathcal{A}^S|\mathcal{E})$, the posterior probability. Using Bayes rule, we can write:

$$P(\mathcal{A}^S|\mathcal{E}) = \frac{P(\mathcal{A}^S)P(\mathcal{E}|\mathcal{A}^S)}{P(\mathcal{E})} \qquad (1)$$

Since the denominator $P(\mathcal{E}|\mathcal{A}^S)$ is 1 and $P(\mathcal{E})$ is a constant for all the assignments we are comparing, it is sufficient to maximize the prior probability $\mathcal{A}^S$, which is much easier to compute (see section 2).

Definition 2 handles the case where the assigned variables are *exactly* statistically independent of the unassigned variables. There remained the problem that modifying the conditional probabilities very slightly would have a major effect on the solution. In fact, if the correlation factor between effects and potential causes is nearly 0, we would also want to conclude that these potential causes are irrelevant. In order to achieve that, we relax the exact independence constraint by requiring that the equality hold only within a factor of $\delta$. That is (evaluating over all possible assignments for a set of variables), if the maximum conditional probability is within a factor of $\delta$ of the minimum conditional probability, then we have $\delta$-independence. Formally:

---

[1] The superscript over the assignment symbol denotes the set of variables assigned by $\mathcal{A}$. If the assignment assigns values to *all* the nodes of $S$, we say that $\mathcal{A}$ is complete w.r.t. $S$.

[2] We use $\uparrow(v)$ to denote the set of immediate predecessors of $v$. We omit the set-intersection operator between sets whenever unambiguous, thus $S\uparrow(v)$ is the intersection of $S$ with the immediate predecessors of $v$.

**Definition 3** *We say that a is δ-independent of b given $\mathcal{A}^S$, where a, b and S are sets of variables (written $In_\delta(a,b|\mathcal{A}^S)$ for short), iff*

$$\min_{\mathcal{A}^b} P(\mathcal{A}^a|\mathcal{A}^S, \mathcal{A}^b) \geq (1-\delta)\max_{\mathcal{A}^b} P(\mathcal{A}^a|\mathcal{A}^S, \mathcal{A}^b) \quad (2)$$

We expand the definition to include the case of $a$ being a (possibly partial) *assignment* rather than a set of variables, by substituting $a$ for $\mathcal{A}^a$ in the above definition. Likewise for the case of $b$ being an assignment. This definition is parametric, i.e. $\delta$ can vary between 0 and 1. We define a $\delta$-independent based assignment as an assignment where each node is $\delta$-independent of its unassigned ancestors given its assigned parents. Formally:

**Definition 4** *An assignment $\mathcal{A}^S$ is δ-independence based iff for every $v \in S$, $In_\delta(\mathcal{A}^{\{v\}}, \uparrow^+(v) - S|\mathcal{A}^{S\uparrow(v)})$.*[3]

The case of $\delta = 0$ reduces to the independence-based assignment criterion. A $\delta$-independence based MAP is defined in the same way as independence-based MAPs, using $In_\delta$ in place of $In$.

### 1.2 BEST-FIRST MAP ALGORITHM

We presented an algorithm for finding complete (rather than partial) MAP assignments to belief networks in [Shimony and Charniak, 1990]. The algorithm finds MAP assignments in linear time for belief networks that are polytrees (when appropriate bookkeeping, not discussed here, is used). The algorithm is potentially exponential time in the general case, as the problem is provably NP-hard.

An agenda of states is kept (or assignments), sorted by current probability, which is a product of all conditional probabilities seen in the current expansion. The operation of the algorithm is shown in the figure 1. An agenda item is *complete* iff all the variables are assigned. Expansion consists of selecting a fringe node (i.e. a node that has unassigned neighbors) and creating a new agenda item for each of the possible assignments to neighboring nodes. The heuristic evaluation function for an agenda item, which is an assignment $\mathcal{A}^S$ to the set of nodes $S$, is the following product:

$$H(\mathcal{A}^S) = \prod_{v \in G(S)} P(\mathcal{A}^{\{v\}}|\mathcal{A}^{\uparrow(v)}) \quad (3)$$

where $G(S) = \{v|v \in S \land \forall w \in \uparrow(v), w \in S\}$, i.e. the product is over all assigned nodes which have all their parents assigned as well. Clearly the evaluation function is precise for complete assignments, as the product reduces to exactly the joint distribution of the network in that case. $H$ is also optimistic, because if some nodes are not assigned, it essentially assumes that their probability is 1. Thus, the evaluation function $H$ is heuristically admissible.

---
[3] We use $\uparrow^+$ to denote the non-reflexive transitive closure of $\uparrow$. Thus $\uparrow^+(v)$ is the set of ancestors of $v$.

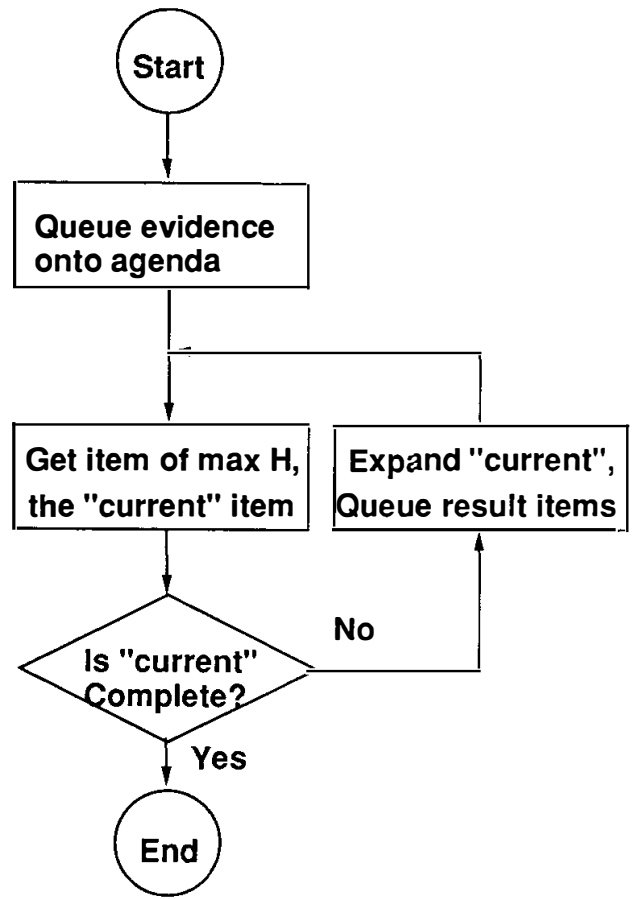

Figure 1: Top Level of Algorithm for Finding MAPs

The advantage of this best-first algorithm is that it can be easily modified to produce the next-best complete assignments in order of decreasing probability. This is done in the following manner (see figure 1): instead of ending with the first complete assignment, output it, and simply continue to loop (getting the next agenda item).

In the following sections, we discuss properties of independence-based and $\delta$-independence-based partial assignments that allow us to use essentially the same algorithm (with only local modifications) to compute them. We then present the modifications required to produce the IB-MAP algorithm. A formal specification of the IB-MAP algorithm and a proof of its correctness follows. We conclude with suggestions of how to modify the algorithm to find $\delta$-independence based MAPs.

## 2 IB-MAP ALGORITHM

We begin by informally introducing the changes required to convert the complete MAP algorithm to an IB-MAP algorithm. We then proceed to define the terms and the algorithm formally, and prove its correctness.

### 2.1 ALGORITHM MODIFICATIONS

The algorithm modifications needed to compute the independence-based partial MAP are in checking whether an agenda item is complete, and in the expansion of an agenda item. Completeness checking in the modified algorithm is different in that an agenda item may be complete even if not all variables are assigned. Specifically, an agenda item is complete iff it is an independence-based (possibly partial) assignment. The other conditions for the agenda item being an IB-MAP are guaranteed because the evidence nodes are assigned initially. Checking whether an assignment is independence-based is easy, due to the following theorem (the locality theorem):

**Theorem 1** *If $\mathcal{A}^S$ is a complete assignment to all the nodes of subset $S$ of a belief network $B$, and for every node $v \in S$, $In(\mathcal{A}^{\{v\}}, \uparrow(v) - S | \mathcal{A}^{S\uparrow(v)})$, then $\mathcal{A}^S$ is an independence-based partial assignment to $B$.*

The claim is essentially that if conditional independence holds locally (i.e. with respect to just the immediate predecessors, as opposed to *all* ancestors, as in the definition of independence-based partial assignments), then it also holds globally. The theorem allows us to test whether an assignment is independence-based in time linear in the size of the network, and is thus an important theorem to use when we are considering the development of an algorithm to compute independence-based partial MAPs. The following theorem allows us to compute $P(\mathcal{A}^S)$ easily:

**Theorem 2** *If $In(v, \uparrow(v) - S | \mathcal{A}^{S\uparrow(v)})$ holds for every node $v \in S$, then the probability of the assignment is:*

$$P(\mathcal{A}^S) = \prod_{v \in S} P(\mathcal{A}^{\{v\}} | \mathcal{A}^{S\uparrow(v)}) \quad (4)$$

Theorem 2 allows us to calculate $P(\mathcal{A}^S)$ in linear time for independence-based partial assignments, as the terms of the product are simply conditional probabilities that can be read off from the conditional distribution array (or any other representation) of nodes given their parents.

Another modification is required because, when extending a node, we may want to leave some of the parents unassigned, as we will show presently. Also, only nodes with unassigned *parents* are considered fringe nodes, since we do not need to assign nodes with no evidence nodes below them.

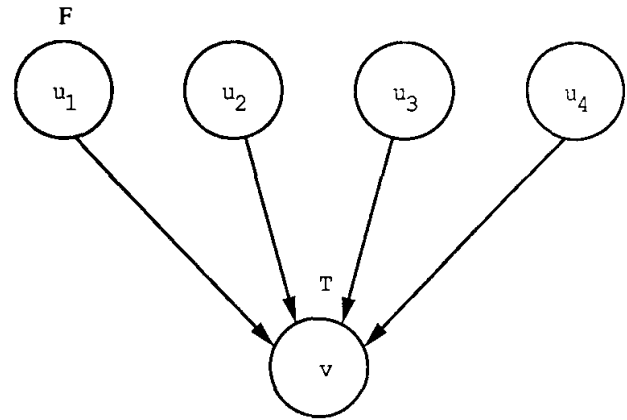

P(v| some parent true) = 0.9

P(v| all parents false) = 0.1

Figure 2: Expanding a Node

To take advantage of theorem 1, we precompute for each node $v$ a set of all the cases where partial independence occurs. We do that in the following manner. The space defined by an assignment to a node $v$ and some of its parents, (where the other parents are not assigned), defines a hypercube $\mathcal{H}$ of possible value assignments. If the probability of $v$ is the same given any assignment in $\mathcal{H}$, then $\mathcal{H}$ is an *independence-based hypercube*. Another way to look at this is: an independence-based hypercube is a sub-space of the conditional distribution array (of $v$ given its parents) with equal conditional probability entries. Consider, for example the "leaky" OR node $v$ of figure 2, where $P(v = T | u_i = T) = 0.9$ for $1 \le i \le 4$ is independent of $u_j$, $j \ne i$. This defines four 3-dimensional hypercubes of "don't-care" values. We also have the 1-dimensional hypercube where all $u_i = F$. When the algorithm expands $v$, it only assigns values to parents of which $v$ is not independent (given the assignment to its other parents), i.e. it generates one agenda item for each of the above independence-based hypercubes.

Naturally, since a belief net is not always a tree, some nodes may already be assigned. Consider, for example, figure 2. We are at the OR node $v$, with parents $u_1$, $u_2$, $u_3$, $u_4$, where $v$ has the value $T$, and $u_1$ has already been assigned $F$. We now have to expand all the states of the nodes $u_i$, the parents of $v$. We would, however, like to generate as few new assignments as possible, while guaranteeing that the IB-MAP is still reachable.

In the complete MAP case, we add the following 8 assignments for the nodes $(u_2, u_3, u_4)$:

$$\{(F, F, F), (F, F, T), (F, T, F), (F, T, T),$$
$$(T, F, F), (T, F, T), (T, T, F), (T, T, T)\}$$

That is, all possible complete assignments to these three variables. When we need to find the partial

MAP, however, only the following 4 assignments are added:

$$\{(T,U,U), (U,T,U,), (U,U,T), (F,F,F)\}$$

where $U$ stands for "unassigned". If a hypercube is ruled out by a prior assignment to a parent node (as is the case $u_1 = T$ here), it is ignored. Otherwise, the hypercubes are unified with the prior assignment, as in this case, the 3-dimensional hypercubes are reduced to 2-dimensional hypercubes by the prior assignment of $u_1 = F$. All the other assignments are redundant, because they would assign values to variables that cannot change the probability of $v$, and are subsumed by the 4 assignments listed above.

Finally, to compute next-best partial assignments in decreasing order, we perform the same simple modification as for the complete MAP algorithm: simply continue to run, producing independence based partial assignments. A useful termination condition is now a probability threshold, i.e. stop producing assignments once the probability of an assignment is below some fraction of that of the first partial MAP produced.

## 2.2 FORMAL DEFINITION OF THE ALGORITHM

We define the algorithm in terms of an input assignment $\mathcal{E}$, the evidence, and and output IB assignment. We shall define an expansion operator $\tau$, and a termination condition, and show that the algorithm terminates with an IB-MAP.

We assume a total ordering $\mathcal{O}$ on the nodes, such that no node comes before its (possibly indirect) descendents. That is always possible, because belief networks are directed acyclic graphs (DAGs). A fringe node $w$ is minimal in an assignment if it is the first node w.r.t. the ordering $\mathcal{O}$ that has unassigned parents. If $w$ is a fringe node in an assignment, such that the independence-based assignment condition holds at $w$ w.r.t. the assignment, then it is an independence-based inactive (or just inactive, for short) fringe node. If the latter does not hold, then it is an active fringe node. If $w$ is the first active node in the assignment, it is called a minimal active fringe node. Given an assignment and an ordering, the minimal active fringe node is unique. Unless otherwise specified, we shall assume that an implicit ordering $\mathcal{O}$ is present, and define the function $index : nodes(B) \to \mathcal{N}$, the index of a node w.r.t. $\mathcal{O}$.

An assignment $\mathcal{A}^{\{w\}\cup X}$ to a node and a subset of its parents $(X \subseteq \uparrow (w))$ is called a *hypercube* based on $w$. If $\mathcal{A}^{\{w\}\cup X}$ is complete w.r.t. $w$ and $X$ and $P(\mathcal{A}^{\{w\}}|\mathcal{A}^X)$ is independent of the nodes $\uparrow(w) - X$, that is[4]:

$$\exists p \ \forall \mathcal{A}^V \in \mathcal{A}_c^{\uparrow(w)-X} \quad P(\mathcal{A}^{\{w\}}|\mathcal{A}^X, \mathcal{A}^V) = p \quad (5)$$

---
[4] $\mathcal{A}_c^{\uparrow(w)-X}$ is the set of all complete assignments to the nodes $\uparrow(w) - X$

then $\mathcal{A}^{\{w\}\cup\uparrow(w)}$ is an independence-based hypercube (acronym IB hypercube), and $p$ is the conditional probability of the hypercube.

**Definition 5** *An IB hypercube $\mathcal{A}^{\{w\}\cup X}$ based on $w$ is maximal if there does not exist a different independence-based hypercube $\mathcal{B}^{\{w\}\cup Y}$ based on $w$ that subsumes it (i.e. it is maximal with respect to subsumption).*

The maximal IB hypercube based on $w$ is not always unique. Note also that a maximal IB hypercube has the *smallest* set of nodes assigned. We currently assume, for computation of hypercubes, that the distribution is strictly positive.

**Theorem 3** *If independence-based assignment $\mathcal{A}^S$ is subsumed by the evidence $\mathcal{E}$, but is not evidentially supported w.r.t. $\mathcal{E}$, then there exists an independence-based assignment $\mathcal{A}^{S'}$ that subsumes $\mathcal{A}^S$ and is evidentially supported w.r.t. $\mathcal{E}$.*

Proof: By construction: we show that we can drop all the nodes that have no evidence nodes below them from the assignment $\mathcal{A}^S$. Since the belief network structure is a DAG, then so is any subgraph. Order nodes of $S$ that are not ancestors of some node in $E$ (nodes in $E$ are considered to be ancestors here) in a list such that no node precedes its descendents. Now, proceed to eliminate nodes from the list (and from the assignment), in order of the elements of $S$. As each node is eliminated, the assignment remains independence-based, as only nodes with no children are eliminated, and the independence-based assignment criterion for each node depends only on ancestor nodes. We can thus eliminate the entire list, and remain with an assignment that is evidentially supported, is still subsumed by $\mathcal{E}$, and is independence-based. Q.E.D.

**Theorem 4** *If $\mathcal{A}^S$ is an independence-based assignment that is subsumed by $\mathcal{E}$, then there exists an independence-based assignment $\mathcal{A}^{S'}$ that subsumes $\mathcal{A}^S$ and is properly evidentially supported w.r.t. $\mathcal{E}$.*

Proof: By construction: we show that we can delete from the assignment $\mathcal{A}^S$, all the nodes that have no evidence nodes below them, as well as all nodes for which no path to an evidence node (that traverses only nodes in $S$) exists. Remove from the assignment $\mathcal{A}^S$ all nodes that are not ancestors of $E$ as in the proof of theorem 3. Then, remove all the nodes $T$ that have no path to a node in $E$ that lies entirely in $S$, in a similar manner: sort the nodes of $T$ into a list such that no node precedes its descendents. Removing the nodes of $T$ will achieve a properly evidentially supported assignment, if we preserve the independence-based assignment condition. But removing the nodes of $T$ in sequence will always preserve the criterion, because no node $v$ is removed if it has children in the resulting assignment (if it did, then the node $v$ would not have been in $T$, as there would be a path from $v$ to a node in $E$). Q.E.D. Theorems 3 and 4 are further support for our intuition for requiring that IB-MAPs be properly evidentially supported.

Let $\mathcal{A}_p$ be the set of all possible (either partial or complete) assignments. We define our expansion operator $\tau : \mathcal{A}_p \cup 2^{\mathcal{A}_p} \to 2^{\mathcal{A}_p}$, as follows:

**Definition 6** $\tau(S)$ *consists of exactly the assignments* $\mathcal{A}^S$ *that obey the following conditions:*

- *If* $S \in \mathcal{A}_p$, *then* $S$ *subsumes* $\mathcal{A}^S$ *and there exists a fringe node* $w \in S$ *and a maximal IB hypercube* $B^{\{w\} \cup X}$ *(based on* $w$, *where exactly the nodes* $X \subseteq \uparrow(w)$ *are assigned), such that both the following conditions hold:*

  1. $S = nodes(S) \cup X$
  2. $\mathcal{A}^S = S \cup H^{\{w\} \cup X}$

- *If* $S \in 2^{\mathcal{A}_p}$, *then exists an assignment* $\mathcal{A}^{S'} \in S$ *that subsumes* $\mathcal{A}^S$, *such that exist a fringe node* $w \in S'$ *and a maximal IB hypercube* $B^{\{w\} \cup X}$ *(based on* $w$, *where exactly the nodes* $X \subseteq \uparrow(w)$ *are assigned), such that both the following conditions hold:*

  1. $S = S' \cup X$
  2. $\mathcal{A}^S = \mathcal{A}^{S'} \cup H^{\{w\} \cup X}$

*In both cases, the hypercube should not contradict the assignment to variables that are already assigned.*

We say that $w$ is the fringe node expanded by $\tau$. Essentially, applying the $\tau$ operator is equivalent to taking all the IB-hypercubes at a certain node $w$, and creating a new assignment for each hypercube. The new assignment is a union of the old assignment and the hypercube.

**Theorem 5** *If assignment* $\mathcal{A}^S$ *is* $\tau$-*reachable from* $\mathcal{E}$, *then it is properly evidentially supported by* $\mathcal{E}$.

Proof: By induction on the number of applications of the *tau* operator. The theorem clearly holds for 0 applications, as the only assignment in that case is $\{\mathcal{E}\}$, which is clearly properly evidentially supported. Now, assuming that the theorem holds for $n$ applications of $\tau$, then another application of $\tau$ can only assign values to nodes that are in some IB-hypercube based on a node $w$ already assigned. IB-hypercubes assign only direct parents of $w$, and $w$ is either in $E$ or there exists a path from $w$ to $E$ passing only through assigned nodes, by the induction hypothesis. Hence, there will always by a path from the nodes assigned in the $n+1$ application of $\tau$ to $E$. The theorem follows by induction. Q.E.D.

An assignment $\mathcal{A}^S$ is IB-*terminated* when each assigned node $w \in S$ either has no parents, or the IB condition holds at $w$. The latter is true iff the assignment for every $w \in S$, $\mathcal{A}^{\{w\} \cup S \uparrow(v)}$ is subsumed by some IB hypercube based on $w$.

**Theorem 6** *Every maximal (w.r.t. subsumption) independence-based assignment* $\mathcal{A}^S$ *that is properly evidentially supported w.r.t.* $\mathcal{E}$ *is* $\tau$-*reachable from* $\mathcal{E}$.

Proof outline: We show that there exists a sequence of assignments $\mathcal{A}^{S'_k}$ of sufficient length, such that each assignment subsumes $\mathcal{A}^S$, $\mathcal{A}^{S'_k} \in \tau(\mathcal{A}^{S'_{k-1}})$, and if $v_k$ is the node expanded by $\tau$, then all the nodes $v_i$, $i \leq k$, are assigned exactly as in $\mathcal{A}^S$. Thus, for some $k$, $\mathcal{A}^{S'_k} = \mathcal{A}^S$. The proof of this theorem shows that it is actually sufficient to expand only the minimal fringe node at each state, rather than all fringe nodes.

Using an agenda $S$ (a set of states, or assignments), evaluation function $H$, evidence $\mathcal{E}$ (where $E = nodes(\mathcal{E})$) and expansion operator $\tau$, the algorithm is defined formally as follows:

1. Set $S = \{\mathcal{E}\}$, and $i = min_{v \in E}\ index(v)$.
2. Set $\mathcal{A}^S$ to be a member of $S$ of maximum $H(\mathcal{A}^S)$, and remove it from $S$.
3. If $\mathcal{A}^S$, is IB-terminated, halt ($\mathcal{A}^S$ is an IB-MAP).
4. Set $S = (S \cup \tau(\mathcal{A}^S)) - \mathcal{A}^S$, and go to step 2.

The evaluation function $H$ is similar the one for the complete MAP algorithm. The only difference is that the (conditional) probability of a node $v$ is included in the product if the IB condition holds at $v$, as well as when all its parents are assigned:

$$H(\mathcal{A}^S) = \prod_{v \in G(S)} P(\mathcal{A}^{\{v\}} | \mathcal{A}^{\uparrow(v)}) \quad (6)$$

$$G(S) = \{v | v \in S \land In(\mathcal{A}^{\{v\}}, \uparrow(v) - S | \mathcal{A}^{S\uparrow(v)})\} \\ \cup \{v | v \in S \land \forall w \in \uparrow(v),\ w \in S\}$$

$H$ is obviously optimistic, and because of theorem 2, it is exact for IB assignments (the goal states). As the algorithm is implemented, $H$ is actually computed before adding an assignment to the agenda, and the agenda is always kept sorted (e.g. using a heap). We now show that the algorithm is correct.

**Theorem 7** *The IB-MAP algorithm terminates, and when it halts it does so with* $\mathcal{A}^S$ *being the most-probable IB assignment that is properly evidentially supported and subsumed by* $\mathcal{E}$.

Proof: The algorithm terminates, because the number of states added to the agenda in step 3 is finite, and since it always adds nodes to each assignment $\mathcal{A}^S$, it will eventually assign all the nodes above $E$, in which case the IB condition is vacuously true. Naturally, the runtime may be exponential. The assignment found when the algorithm terminates is IB (that

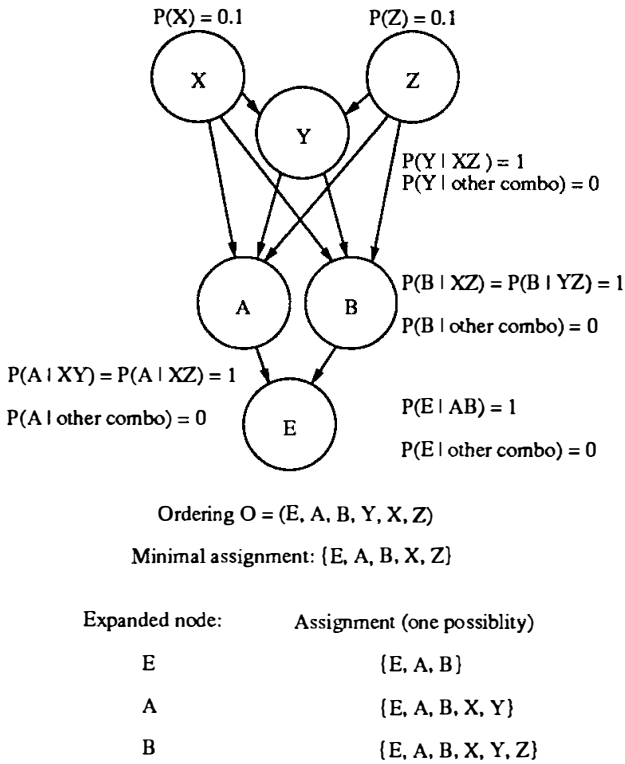

Ordering O = (E, A, B, Y, X, Z)

Minimal assignment: {E, A, B, X, Z}

| Expanded node: | Assignment (one possiblity) |
|---|---|
| E | {E, A, B} |
| A | {E, A, B, X, Y} |
| B | {E, A, B, X, Y, Z} |

Figure 3: How can a non-maximal assignment occur?

is the termination condition). It is properly evidentially supported (from theorem 5) and the fact that all assignments generated are $\tau$ accessible from $\mathcal{E}$. The evaluation function admissible, and all possible maximal properly evidentially supported IB assignments are $\tau$-accessible. The theorem follows from the latter two properties, and from the correctness condition of heuristic search w.r.t. evaluation functions. Q.E.D.

Continuing to run the algorithm after finding a first assignment will find next-best IB-assignments, in decreasing order of probability. Note that theorem 7 does not guarantee a *maximal* (w.r.t. subsumption) IB-MAP. In fact, figure 3 shows a simple counterexample, where all the nodes are binary, $E$ is the evidence node, and is known to be true. Given the set of agenda states shown, the non-maximal assignment, $\{E, A, B, X, Y, Z\}$ is reached. The latter assignment is not maximal w.r.t. subsumption, because the assignment $\{E, A, B, X, Z\}$ subsumes it, and is both IB and properly evidentially supported.

However, for positive distributions, subsumption also implies a higher probability, which guarantees that the IB-MAP found is indeed maximal. For other distributions, to find the maximal IB-MAPs, we need to compare all IB-MAPs with equal probability, which is not hard in most cases. We are assured that the maximal IB-MAP will indeed appear if we continue to run the algorithm, because of theorem 6.

## 3  δ-IB MAP ALGORITHM

In the case of $\delta$-independence based MAPs, we use essentially the same algorithm again, where we need to pre-compute $\delta$-independence hypercubes, rather than independence hypercubes, as in the previous case. However, once that is done, we can again employ local checking:

**Theorem 8** *If $\mathcal{A}^S$ is a complete assignment to all the nodes of subset $S$ of a belief network $B$, and for every node $v \in S$, $In_\delta(\mathcal{A}^{\{v\}}, \uparrow(v) - S | \mathcal{A}^{S\uparrow(v)})$ (for $0 \leq \delta \leq 1$), then $\mathcal{A}^S$ is a $\delta$-independence-based partial assignment to $B$.*

Computing the exact probability of a $\delta$-independence based partial assignment seems to be hard (since we cannot use theorem 2, and would need to find posterior probabilities of non-root nodes), but the following easily computable bound inequalities are always true:

$$P(\mathcal{A}^S) \leq \prod_{v \in S} \max_{\mathcal{A}^{U\uparrow(v)}} P(\mathcal{A}^{\{v\}} | \mathcal{A}^S, \mathcal{A}^{U\uparrow(v)}) \quad (7)$$

$$P(\mathcal{A}^S) \geq \prod_{v \in S} \min_{\mathcal{A}^{U\uparrow(v)}} P(\mathcal{A}^{\{v\}} | \mathcal{A}^S, \mathcal{A}^{U\uparrow(v)})$$

These bounds get better as $\delta$ approaches 0, as their ratio is at least $(1 - \delta)^{|S|}$.

Since getting the exact probability is hard, but the upper and lower bounds above (denoted $\mathcal{U}(\mathcal{A})^S$ and $\mathcal{L}(\mathcal{A})^S$ respectively) are easily computable, a post-processing step may be needed, to select the most-probable assignment from a number of possible candidates. During the first part of the algorithm, the assignments are sorted in the agenda according to $\mathcal{U}(\mathcal{A})$. We need to collect the *set* of assignments $F$ such that for all assignments $\mathcal{A}$ not in $F$, $\mathcal{U}(\mathcal{A})$ is smaller than $\mathcal{L}(\mathcal{A})'$, for *some* $\mathcal{A}'$ in $F$. This assures us that the most-probable assignment is indeed in $F$. Hopefully, $F$ is a small set (as indeed it will prove to be in almost all cases where one explanation clearly stands out). Then, we evaluate the exact probability of the assignments in $F$ in parallel by adding AND nodes for all of them and evaluating the diagram exactly once. Naturally, if $F$ happens to contain only one assignment, we do not need to post-process the results.

## 4  FUTURE WORK

The locality property of IB assignments and $\delta$-IB assignments, i.e. the fact that local testing is sufficient to determine whether an assignment is an IB assignment, together with a quick way of computing its probability, makes it possible, in principle, to use other types of algorithm. Future research will determine whether random simulation techniques will prove useful.

Another possibility is to reduce IB-MAP computation to complete MAP computation (on a different belief network). That will allow us to use any complete MAP algorithm, such as belief updating ([Pearl, 1988]), to be used. This may prove to be faster than our algorithms for networks with a small maximal clique size (for using clustering), or a small cutset size (for using conditioning). The algorithm presented in [Santos Jr., 1991a] and [Santos Jr., 1991b] may also prove useful for our purposes, if it can be easily extended to work on multiple-valued nodes.

## 5 SUMMARY

We introduced irrelevance-based partial MAPs in order to solve the overspecification problem inherent in complete MAP explanation. We defined two methods for defining what nodes are irrelevant, one based on exact independence, the other based on approximate independence. We then discussed properties of the resulting partial MAPs that allowed us to transform an existing best-first search for complete MAPs into one that computes irrelevance-based partial MAPs. The modifications required were minimal.

For independence-based assignments, we showed that we can check whether an assignment is independence based using only local information, (i.e. they are "locally recognizable"). Computation of the probability of such an assignment is also easy, and can be done using only $|S|$ conditional probability array entries. We showed how to adapt our best-first MAP algorithm to compute IB-MAPs, and proved the correctness of the resulting algorithm.

$\delta$-independence-based assignments were shown to be locally recognizable. Computing the exact probability is hard, but good, easily computable bounds are available. It is possible to compute the exact probabilities in parallel, using only one belief-network evaluation (with extra nodes).

We have implemented algorithms for finding IB-MAPs and for finding $\delta$IB-MAPs, and their running time seems roughly comparable to the running time of our complete MAP algorithm running on the same networks. That is not surprising, as the additional work required for expansion and for completion testing in small, and the number of states expanded is usually smaller than for complete MAPs.

### Acknowledgements

This work has been supported in part by the National Science Foundation under grants IST 8416034 and IST 8515005 and Office of Naval Research under grant N00014-79-C-0529. The author is funded by a Corinna Borden Keen Fellowship. Special thanks to Eugene Charniak for helpful suggestions and for reviewing drafts of the paper.